\pdfoutput=1

\documentclass[11pt]{article}

\usepackage[final]{acl}

\usepackage{times}
\usepackage{latexsym}

\usepackage[T1]{fontenc}

\usepackage[utf8]{inputenc}

\usepackage{microtype}

\usepackage{inconsolata}

\usepackage{graphicx}

\definecolor{hycolor}{rgb}{0.7,0.7,0.3} 

\title{A Semantic-Aware Layer-Freezing Approach to\\Computation-Efficient Fine-Tuning of Language Models}

\author{Jian Gu \\
  Monash University \\
  \texttt{jian.gu@monash.edu} \\\And
  Aldeida Aleti \\
  Monash University \\
  \texttt{aldeida.aleti@monash.edu} \\\AND
  Chunyang Chen \\
  Technical University of Munich \\
  \texttt{chun-yang.chen@tum.de} \\\And
  Hongyu Zhang \\
  Chongqing University \\
  \texttt{hyzhang@cqu.edu.cn} \\
}

\usepackage{mystyle}

\DeclareAcronym{api}{
	short = API,
	long = {Application Program Interface}
}

\DeclareAcronym{awgn}{
	short = AWGN,
	long = {additive white Gaussian noise}
}

\DeclareAcronym{vae}{
	short = VAE,
	long = {Variational AutoEncoder}
}

\DeclareAcronym{bert}{
	short = BERT,
	long = {Bidirectional Encoder Representations from Transformers}
}

\DeclareAcronym{roberta}{
	short = RoBERTa,
	long = {Robustly optimized BERT approach}
}

\DeclareAcronym{ast}{
	short = AST,
	long = {Abstract Syntax Tree}
}

\DeclareAcronym{bpe}{
	short = BPE,
	long = {Byte-Pair Encoding}
}

\DeclareAcronym{cfg}{
	short = CFG,
	long = {Control Flow Graph}
}

\DeclareAcronym{dcg}{
	short = DCG,
	long = {Discounted Cumulative Gain}
}

\DeclareAcronym{gpt}{
	short = GPT,
	long = {Generative Pretrained Transformer}
}

\DeclareAcronym{ir}{
	short = IR,
	long = {Information Retrieval}
}

\DeclareAcronym{lstm}{
	short = LSTM,
	long = {Long Short-Term Memory}
}

\DeclareAcronym{clm}{
	short = CLM,
	long = {Casual Language Modeling}
}

\DeclareAcronym{mlm}{
	short = MLM,
	long = {Masked Language Modeling}
}

\DeclareAcronym{mem}{
	short = MEM,
	long = {Multimodal Embedding Model}
}

\DeclareAcronym{cp}{
	short = CP,
	long = {Continuous Pretraining}
}

\DeclareAcronym{if}{
	short = IF,
	long = {Intermediate Finetuning}
}

\DeclareAcronym{mmpf}{
	short = MMPF,
	long = {Massive Multitask Pre-Finetuning}
}

\DeclareAcronym{aif}{
	short = AIF,
	long = {Adaptive Intermediate Finetuning}
}

\DeclareAcronym{mrr}{
	short = MRR,
	long = {Mean Reciprocal Rank}
}

\DeclareAcronym{ndcg}{
	short = NDCG,
	long = {Normalized Discounted Cumulative Gain}
}

\DeclareAcronym{nlp}{
	short = NLP,
	long = {Natural Language Processing}
}

\DeclareAcronym{nlp_pt}{
	short = NLP\textsubscript{PT},
	long = {Next Line Prediction}
}

\DeclareAcronym{nmt}{
	short = NMT,
	long = {Neural Machine Translation}
}

\DeclareAcronym{nsp}{
	short = NSP,
	long = {Next Sentence Prediction}
}

\DeclareAcronym{rnn}{
	short = RNN,
	long = {Recurrent Neural Network}
}

\DeclareAcronym{cnn}{
	short = CNN,
	long = {Convolutional Neural Network}
}

\DeclareAcronym{tf-idf}{
	short = tf-idf,
	long = {term frequency–-inverse document frequency}
}

\DeclareAcronym{anova}{
	short = ANOVA,
	long = {ANalysis Of VAriance}
}

\DeclareAcronym{da}{
	short = DA,
	long = {Domain-Adaptive}
}

\DeclareAcronym{ta}{
	short = TA,
	long = {Task-Adaptive}
}

\DeclareAcronym{ma}{
	short = MA,
	long = {Multiphase Adaptive}
}

\DeclareAcronym{ca}{
	short = CA,
	long = {Concept Annotation}
}

\DeclareAcronym{ce}{
	short = CE,
	long = {Concept Extrapolation}
}

\DeclareAcronym{ci}{
	short = CI,
	long = {Concept Interpolation}
}

\DeclareAcronym{gru}{
	short = GRU,
	long = {Gated Recurrent Unit}
}

\DeclareAcronym{sota}{
	short = SOTA,
	long = {state-of-the-art}
}

\DeclareAcronym{lcs}{
	short = LCS,
	long = {Longest Common Sequences}
}

\begin{document}
\maketitle
\begin{abstract}
Finetuning language models (LMs) is crucial for adapting the models to downstream data and tasks. However, full finetuning is usually costly. Existing work, such as parameter-efficient finetuning (PEFT), often focuses on \textit{how to finetune} but neglects the issue of \textit{where to finetune}. As a pioneering work on reducing the cost of backpropagation (at the layer level) by answering where to finetune, we conduct a semantic analysis of the LM inference process. We first propose using transition traces of the latent representation to compute deviations (or loss). Then, using a derived formula of scaling law, we estimate the gain of each layer in reducing deviation (or loss). Further, we narrow down the scope for finetuning, and also, study the cost-benefit balance of LM finetuning. 
We perform extensive experiments across well-known LMs and datasets. The results show that our approach is effective and efficient, and outperforms the existing baselines.
Our approach is orthogonal to other techniques for improving finetuning efficiency, such as PEFT methods, offering practical values on LM finetuning. 
\end{abstract}

\section{Introduction}
\label{sec:introduction}

With the rapid advancements and notable performance of language models, their application has extended to numerous downstream tasks~\cite{Bommasani2021OnTO}.
Fine-tuning techniques are pivotal in augmenting the capabilities of language models~\cite{Raffel2019ExploringTL,Ouyang2022TrainingLM}.
For example, \textsc{Code Llama} is a code-specialized LM and is finetuned on 100B tokens of Python code for a language-specialized variant~\cite{Touvron2023LLaMAOA,Rozire2023CodeLO}.
The Python variant provides better capabilities in code understanding and generation, since Python is most popular in programming~\cite{carbonnelle_2024,tiobe_2024}.

Compared to their smaller pretrained predecessors, finetuning large LMs offers both advantages and disadvantages. On one hand, the vast number of model parameters triggers the emergent abilities of large LMs~\cite{Wei2022EmergentAO}, leading to superior performance across a variety of tasks, which serves as an excellent foundation for domain-specific finetuning. On the other hand, the extensive parameter size presents challenges for downstream finetuning. For instance, large LMs require greater memory costs and higher computational costs in finetuning.

The challenge is on finding the correlation between performance and efficiency of LM finetuning.
There have been developed techniques such as model quantization and PEFT methods to improve efficiency~\cite{Rokh2022ACS,Han2024ParameterEfficientFF}.
Model quantization reduces the precision of the model and data to reduce the burden of storage and computation. However, the performance of LM finetuning may be damaged to some extent.
Most PEFT methods introduce additional parameters to learn the updates in LM finetuning, and then merge the updates into the LM. Their focus is to reduce the memory cost instead of the computational cost~\cite{Han2024ParameterEfficientFF}.
Overall, there has been relatively little work exploring the correlation between model performance and computational efficiency, that is, whether the performance of LM finetuning can be improved while saving computation cost.
To mitigate the gap, we propose utilizing the semantics in LM latent space to specify the layers that are more in need of finetuning being trainable, and freeze other layers.

Our intuition is that, by interpreting the LM's functionality as a transition of semantics and comparing it with a set of special latent representations, we can estimate the gains of each layer in reducing deviations.
The deviations can be used to evaluate the convergence degree of model layers, and further, as the evidence to decide which layers shall be trainable.
Based on empirical experience and theoretical analysis, the deviations in semantic transitions greatly decide the effects of LM finetuning. By freezing model layers with the maximum gains in reducing deviation and shortening the process of backpropagation, the computation cost may be reduced and meanwhile, the finetuning effects can be improved.
Computational-efficient finetuning via layer-freezing is orthogonal with existing techniques, including model quantization and PEFT methods, so can combine with these techniques to achieve more efficient performance.



In this paper, we realize computation-efficient model finetuning by proposing an effective and reliable layer-freezing approach, referred to as \underline{S}emantic-\underline{A}ware \underline{L}ayer-\underline{F}reezing (\textsc{SALF}).
First, on the shoulder of LM semantics~\cite{gu2024vocabularydefinedsemanticslatentspace,gu2025semetrainingfreelanguagemodel}, we study the phenomenon of semantic transition in LMs. By deriving the scaling law of LM pretraining, we estimate the gains of reducing deviations in each model layer;
Next, our layer-freezing approach finds the model layer whose gain is the maximum and only finetunes the deeper layers;
Last, to support a flexible cost-benefit tradeoff in LM finetuning, we propose a shallow-to-deep policy for layer-freezing under a given budget. We also propose better budget plans for the tradeoff.

We evaluate our approach in fine-tuning diverse datasets on a wide range of modern LMs.
Based on the results, our semantic-based layer-freezing approach performs better than baselines.
Combined with budget plans, our approach can further reduce the computation cost and improve LM performance.
We discuss the insights of efficient finetuning from the perspective of semantics and conclude the findings in LM finetuning.
The replication artifact is available online for open science \footnote{\url{https://github.com/jianguda/salf}}.



Our contributions are as follows:

\smallskip
\noindent

\begin{itemize}
    \item We propose using semantic transition
    to describe the process of LM inference, and the derived formula of scaling law to estimate the capability of model layers, and further study the cost-benefit tradeoff in LM finetuning;
    \item We emphasize the importance of knowing where to finetune, 
    through which we can improve the performance of LM finetuning and save the computation cost. We propose semantic-based layer-freezing as a solution;
    \item We conclude some findings on the behavior of LMs, which can contribute to future work in finetuning and analyzing LMs. Also, we propose planning the budget for a better cost-benefit tradeoff of LM finetuning.
\end{itemize}

\section{Preliminaries}
\label{sec:background}

\subsection{Semantic Field in LM Latent Space}

\begin{figure}[!htb]
    \centering
    \includegraphics[width=1.0\linewidth]{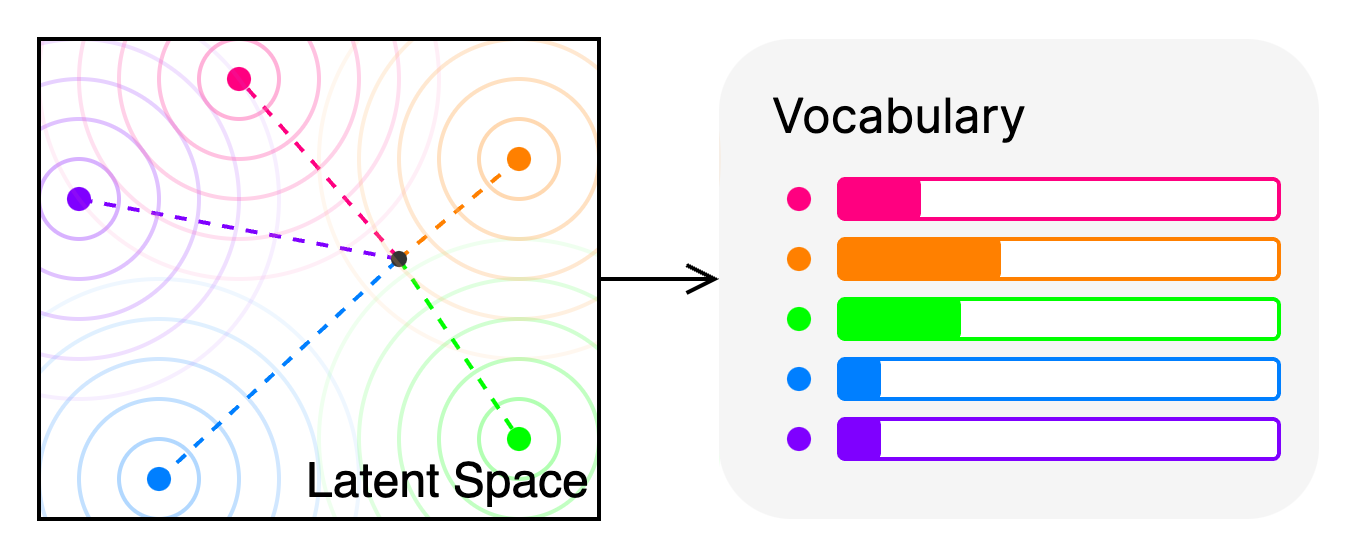}
    \caption{Vocabulary-defined semantics is demonstrated with a LM, whose vocabulary is a collection of colorful labels:
 (1) in the latent space (left), large color dots are the corresponding semantic bases of vocabulary labels. The small dark dot is the hidden state of a given data.
 The similarities of the data with semantic bases are regarded as logits;
 (2) on the vocabulary (right), the logits are normalized as probabilities, and the argmax label is \emph{orange}. Consistently, the nearest semantic basis to the hidden state is the \emph{orange} one.
 }
    \label{fig:logits}
\end{figure}

Based on vocabulary-defined semantics, the semantics of hidden states can be regarded as the overlapping impact of ``semantic fields''~\cite{gu2024vocabularydefinedsemanticslatentspace,gu2025semetrainingfreelanguagemodel}.
The semantic field is similar to the field term in physics, such as electric field, where the electric strength relies on the distance to the center of the field (the electric pole).
The corresponding probabilities on the vocabulary of a representation can be directly computed with its locations in the semantic fields in the latent space, as shown in \cref{fig:logits}.
In contrast, in common practice, the representations in last-layer latent space will undergo a dimensional change to be computed as logits, and then be normalized as the probabilities on the vocabulary.
The dimensional change causes entanglement of semantics, and exacerbates the computation complexity.


The semantics of LM latent space is decided by semantic fields~\cite{gu2024vocabularydefinedsemanticslatentspace,gu2025semetrainingfreelanguagemodel}.
For each label in the vocabulary, there is a corresponding semantic field in the latent space.
The pole of each semantic field is called ``semantic basis'', representing an unmixed and purest meaning.
If the nearest semantic basis to latent representations is the same one, they tend to share the meaning of that semantic basis.
The semantic meaning of a representation in the latent space is decided by the overlapping impact of multiple semantic fields.

The computation of semantic bases is simple.
At the LM input side, we multiply onehot embedding $\vec{e}$ by the embedding matrix $\mathbb{W}_{i}$ to obtain the semantic basis $\vec{r}_{i}=\vec{e}\cdot\mathbb{W}_{i}$.
At the LM output side, due to the opposite operation direction between the embeddings and the representations, we turn to use the pseudoinverse of the LM-head matrix $\mathbb{W}_{o}^+$.
We multiply onehot embedding $\vec{e}$ by the pseudoinverse matrix to obtain the semantic basis $\vec{r}_{o}=\vec{e}\cdot\mathbb{W}_{o}^+$.
Since LM vocabulary is required in the computations, semantic bases only exist in the embedding latent space and the last-layer latent space.

\subsection{Semantic-based Loss Computation}

Based on the local isotropy of LM latent space~\cite{Cai2021IsotropyIT}, the logits in LM training and inference can be computed via similarity measurement (with semantic basis), instead of matrix multiplication (with LM-head matrix). The logits computed in this way is termed as "similarity-based logits"~\cite{gu2024vocabularydefinedsemanticslatentspace}. It proves to have the same effects as the common practice of logits computation, and shows advantages in disentangling the semantics.

\begin{algorithm}
\caption{Semantic Cross-Entropy Loss}
\label{algo:loss-ce}
\begin{algorithmic}

\REQUIRE $N$ semantic bases \texttt{$\vec{b_{i}}$}; ground truth label \texttt{$l$}; last-layer latent repr \texttt{$\vec{r}$}
\ENSURE optimization target \texttt{$loss$}
\STATE \texttt{logits} $\gets$ \texttt{init\_1d\_tensor($N$)}
\FOR{\texttt{$i \gets 0$ \textbf{to} $N$}}
\STATE \texttt{logits[$i$] $\gets$ cosine\_similarity($\vec{r}$, $\vec{b_{i}}$)}
\ENDFOR
\STATE \texttt{loss $\gets$ cross\_entropy\_loss(logits, $l$)}

\end{algorithmic}
\end{algorithm}

In LM finetuning, the logits will be used in loss computation. Taking the cross-entropy loss as an example, we compute the similarities between a given hidden state with semantic bases as similarity-based logits, and then compute with the ground truth for the loss, as shown in \cref{algo:loss-ce}.
In terms of numerical calculations, when computing in the last-layer latent space, it is equivalent to the logits computed via matrix multiplication.

\begin{algorithm}
\caption{Semantic Cosine-Distance Loss}
\label{algo:loss-cos}
\begin{algorithmic}

\REQUIRE $l$-th semantic base \texttt{$\vec{b_{i}}$} (for ground truth label \texttt{$l$}); last-layer latent repr \texttt{$\vec{r}$}
\ENSURE optimization target \texttt{$loss$}
\STATE \texttt{loss $\gets$ 1 - cosine\_similarity($\vec{r}$, $\vec{b_{l}}$)}

\end{algorithmic}
\end{algorithm}

Further, leveraging the disentanglement effects of similarity-based logits, we can compute the loss merely with the corresponding ground truth.
In the loss computation, the hidden state is only computed with one semantic basis solely, instead of with all semantic bases, as shown in \cref{algo:loss-cos}.
In terms of effect, it optimizes the hidden state to make it steer towards the corresponding semantic basis.
The cosine-distance loss is better in computation cost, and its computation shows an intuitive geometric meaning in the latent space.

\section{Computation-Efficient Fine-Tuning}
\label{sec:approach}


\begin{figure}[!htb]
    \centering
    \includegraphics[width=1.0\linewidth]{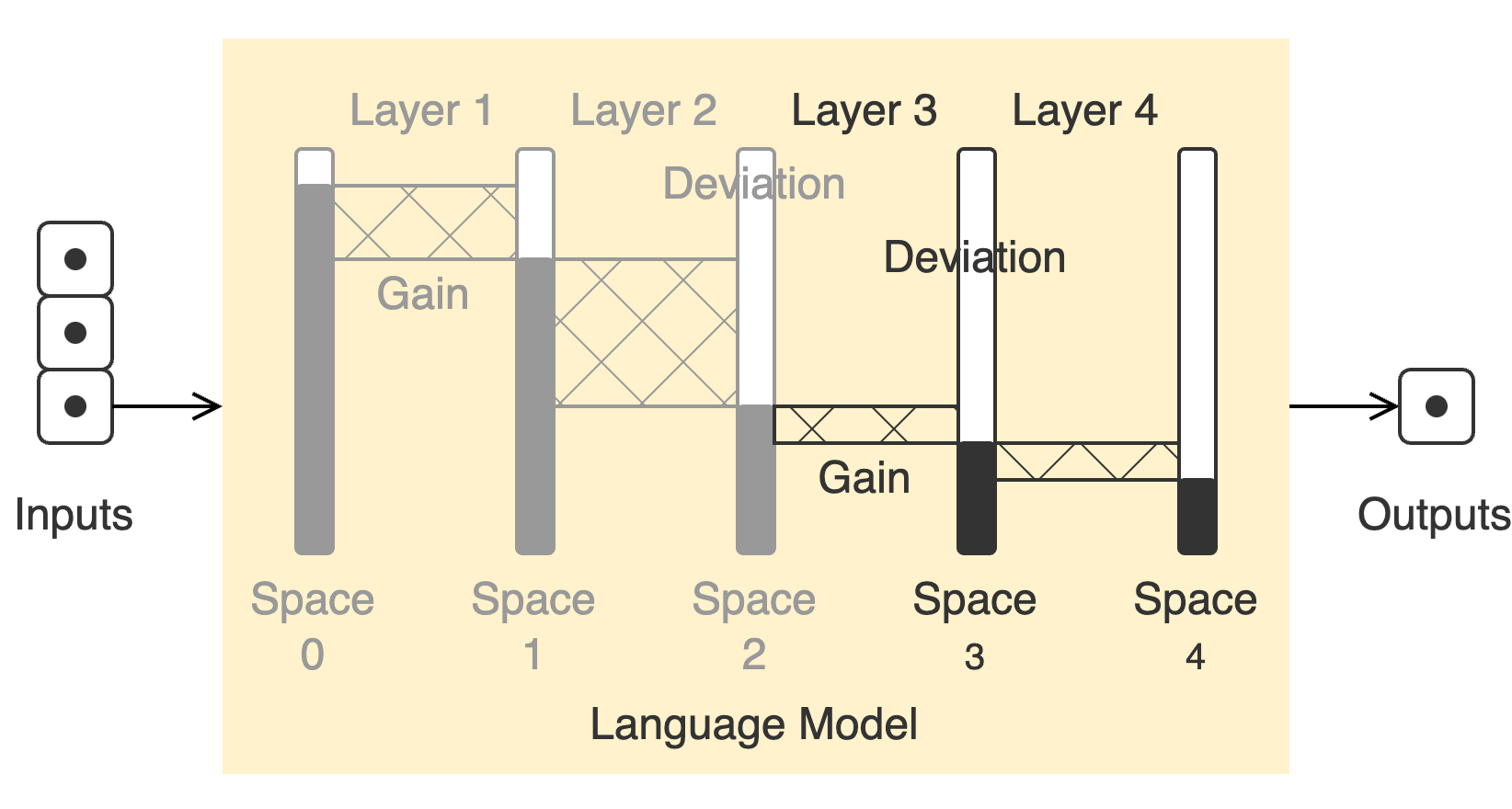}
    \caption{Our SALF approach is demonstrated with a LM with 4 layers (so there are 5 latent spaces). The rectangles with dark bars are the deviations, and the rectangles with cross-hatching are the gains in reducing deviations.
    In LM finetuning, SALF uses a semantic-based analysis to compute the deviations in each latent space, and then uses a derived formula of the scaling law to estimate the gains of each model layer.
    SALF will find the layer with the maximum gain and only finetune the deeper layers. In the illustration, layer 2 is chosen and the first two layers are frozen, so only layer 3 and layer 4 will be finetuned. The layers and spaces marked in gray color mean their gains and deviations will remain unchanged in LM finetuning.}
    \label{fig:salf}
\end{figure}

Semantic-Aware Layer-Freezing, shortened as SALF, is a novel technique to reduce the computation cost in LM finetuning, by freezing certain layers. The core idea is dropping the unnecessary computation in LM backward-pass. Due to the chain rule in loss backpropagation, the computation on shallower layers requires the computation in deeper layers. That is, SALF realizes a computation-efficient LM finetuning by freezing the first few layers.
To guarantee that layer-freezing will not damage the finetuning effects, and even improve the finetuning effects, we proposed a semantic-based analysis on LM inference and a derived formula of scaling law to estimate the convergence of layers. An illustration of our SALF approach is shown in \cref{fig:salf}. We also introduce strategies of assigning data samples under a given budget, to obtain a good cost-benefit balance for layer-freezing.

\subsection{Transitions on Semantics}




\begin{figure*}[!tb]
    \centering
    \includegraphics[width=1.0\linewidth]{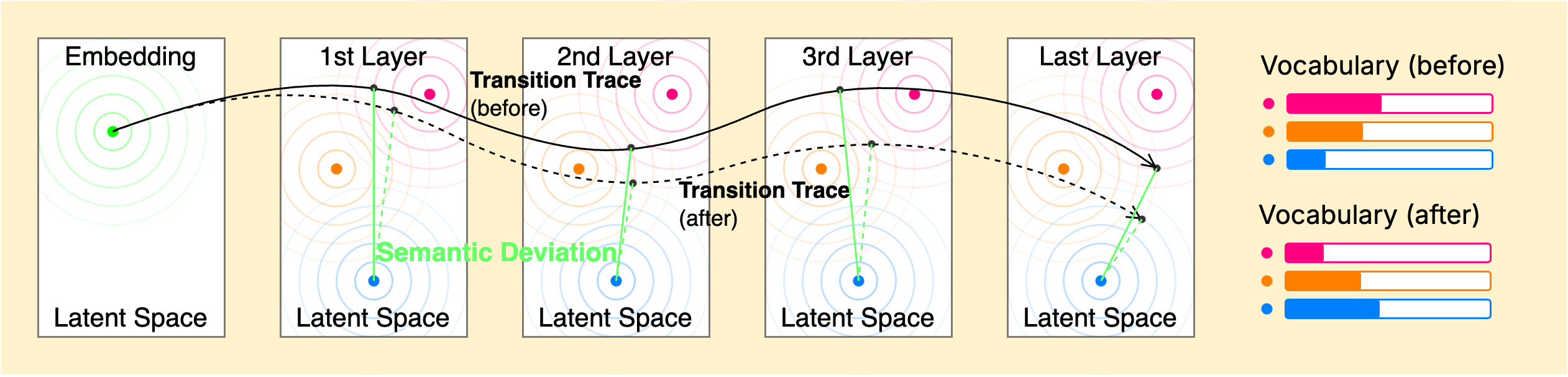}
    \caption{The transition of semantics is illustrated with a 4-layer LM, whose vocabulary is a collection of colorful labels.
The green dot is the medium token (input-side semantic basis) and the blue dot is the ground truth (output-side semantic basis).
The solid/dashed black curves (transition trace) represent the semantic transition of the medium token, defined by the dark dots (hidden states) in each latent space.
The solid/dashed green lines (semantic deviation) indicate the differences between the hidden states and the ground truth, and they differ before and after LM finetuning:
Comparing solid green lines (before finetuning) with dashed green lines (after finetuning), the semantic deviation in each layer is reduced.
Through LM finetuning, the hidden state in the last-layer semantically approaches to the ground truth, and the argmax label becomes from \emph{red} to \emph{blue}.
}
    \label{fig:overview}
\end{figure*}

In next-token prediction, the last token in the given input is used as the medium to compute the next token, denoted as \emph{medium token}.
Influenced by the embeddings of other tokens and the parameters in model layers, the medium token will undergo a layer-by-layer transition on its semantic meaning, denoted as \emph{semantic transition}.
LM finetuning has effects on semantic transition, and the differences before and after finetuning are illustrated in \cref{fig:overview}. We define the involved concepts as below.



\paragraph{Transition Trace}
For a given sequence of $n$ tokens, $t_1, t_2, ..., t_{n}$, assume a $m$-layer LM will predict the next token $t_{n+1}$, the representation of $t_{n}$ undergoes a semantic transition from semantic meaning $i$ to $j$.
The hidden state in each layer is denoted as $f_0, f_1, f_2, ..., f_{m}$ ($f_0$ is the onehot embedding, equals to $i$; while $f_m$ is the last-layer representation, equals to $j$), so the semantic transition defined by these representations is a \emph{transition trace}.




\paragraph{Transition Deviation}
For a semantic transition of a $m$-layer LM, the deviation of the hidden state in the $k$-th layer to the semantic basis of the ground truth, called \emph{semantic deviations}, is denoted as $d_k$. It is measured using cosine similarity, that is, $d_{k} = 1 - \mathtt{cosine\_similarity}(f_{k}, \vec{v})$, where $\vec{v}$ is the semantic basis of ground truth label.
Concerning computation, the semantic deviation is equivalent to the semantic cosine-distance loss, but can also be measured using other metrics.

The semantic deviations before and after finetuning differ. Theoretically and empirically, LM finetuning tends to reduce the deviations.
For a given medium token, in LM finetuning, the transition trace will approach the semantic basis of the corresponding ground truth. The approach will be reflected in the deviation in each layer.
By probing the situation of each layer, the semantic deviation will be reduced as well. That means, the hidden state will approach the semantic basis of the corresponding ground truth.

Further, semantic deviations can be regarded as the evaluation metrics of the capability of model layers.
In the latent space of the LM last-layer, the representation of the medium token is intended to be close enough to the semantic basis (namely the ground truth).
If the hidden states in the middle layers are close to the semantic basis of the corresponding ground truth, then the hidden state in the last layer is likely to be close to the semantic basis as well.
Therefore, leveraging the semantic deviations, model layers can be finetuned selectively.


\subsection{Layer-level Convergence Estimation}

We propose an intuitive method to measure the performance of each model layer leveraging scaling laws.
Scaling laws refer to empirical relationships that describe how the model performance improves with increasing resources, including data amount, model size, and convergence degree (which is often revealed as computational power).

According to the compute-optimal scaling law of LM pretraining~\cite{Hoffmann2022TrainingCL,Zhang2024WhenSM}, the training loss follows a parametric function of the information entropy of training data $E$, the number of model parameters $N$ and the amount of data tokens $D$. The function is shown as \cref{eq:law_training}.
In terms of definition, the second term $\frac{a}{N^\alpha}$ is the ideal capability of model, and the third term $\frac{b}{D^\beta}$ is the finite optimization of data.

\begin{equation}
\label{eq:law_training}
\hat{L}_{pretrain} \triangleq {E} + \frac{a}{N^\alpha} + \frac{b}{D^\beta}
\end{equation}

By performing a slight derivation on the definition of the scaling law, the relationships between the finetuning loss (of the data for finetuning) and resources (used in LM pretraining) can be described as shown in \cref{eq:law_inference}.
The first term ${L}_{0}(E')$ is the loss when interpreting the information entropy of the given data. It only depends on the embedding process and the LM-head, excluding all model layers, since the information entropy is converted to predictable tokens by LM vocabulary.
The second term ${C}(N, D)$ is the capability of models in finetuning, which is the degree close to the convergence. It is a function of data amount $D$ and model size $N$, corresponding to the latter two terms in \cref{eq:law_training}.
In the derived formula, the first term remains stable, while the second term will be larger as the finetuning goes on. It indicates an improved convergence, leading to a smaller loss.

\begin{equation}
\label{eq:law_inference}
\hat{L}_{finetune} \triangleq {L}_{0}(E') - {C}(N, D)
\end{equation}

By targeting a given model, the capability of different layers can be estimated and compared.
For a $m$-layer LM, where layers are denoted as $l_1, l_2, ..., l_{m}$, we define ``virtual submodel'' as the truncated models starting from the shallowest layer. The $k$-th virtual submodel, denoted as $v_{k}$, is composed of $l_1, l_2, ..., l_{k}$ (as well as the embedding layer and the LM-head).
Meanwhile, the loss of all $m$ virtual submodels are computed in one-time of LM forward-pass, so the capability of $v_{k}$ are computed as ${C_{v_k}} \triangleq L_{0} - L_{k} $.
Further, we can compare the capability of model layers. The loss gain of $l_{k}$, denoted as ${G_{l_k}}$, indicates the capability difference between $v_{k}$ and $v_{k-1}$, so we have ${G_{l_k}} \propto {C_{v_k}} - {C_{v_{k-1}}}$. Since $L_{0}$ the remains same, the loss gain are reduced as ${G_{l_k}} \propto {L}_{k-1} - {L}_{k}$.
When the gain of $l_{k}$ is positive, the capability of $v_{k}$ is better than $v_{k-1}$. A larger gain indicates stronger improvement between neighboring submodels.

\subsection{Semantic-Aware Layer-Freezing}

In next-token prediction, via LM finetuning, the last-layer hidden state of the medium token is close enough to the ground truth.
The finetuning process can be explained as divide-and-conquer: If the hidden state is closer to the virtual one in the $k$-th layer, then they tend to be closer as well in the $(k+1)$-th layer.
It is consistent with an existing empirical finding that, when viewed through the output lens, the hidden states across LM layers yield distributions that converge monotonically to the final prediction~\cite{Belrose2023ElicitingLP}.
By making the hidden state close enough to the semantic basis of the ground truth in each layer, the representation in the last-layer tends to be close to the ground truth as well.

Based on the explanation, we propose a layer-freezing method to accelerate finetuning. The idea is simple: \emph{instead of finetuning from the first-layer, we find the layer where the gain is the largest and then finetune from there to the last-layer}.
We call the layer having the largest gain as the end-of-freezing layer, short as \emph{eof-layer}.
The shallow layers will be frozen so only the eof-layer and deeper layers are trainable, as shown in \cref{algo:salf}.


\begin{algorithm}
\caption{Semantic-Aware Layer-Freezing}
\label{algo:salf}
\begin{algorithmic}[1]

\REQUIRE \texttt{model}, \texttt{datum}

\STATE \emph{\# (a) compute deviations of latent spaces}
\STATE \texttt{deviations} $\gets$ \texttt{empty list}
\STATE \texttt{latent\_reprs} $\gets$ \texttt{model(datum)}
\STATE \texttt{semantic\_bases} $\gets$ \texttt{VDS(model)}
\FOR{\texttt{id} $\gets$ 0 \textbf{to} \texttt{layer\_num+1}}
    \STATE \texttt{deviation} $\gets$ \texttt{compute\_deviation(\\~~\texttt{latent\_reprs[id]},\texttt{semantic\_bases})}
    \STATE \texttt{deviations.add(deviation)}
\ENDFOR
\STATE \emph{\# (b) compute gains of model layers}
\STATE \texttt{layer\_gains} $\gets$ \texttt{empty list}
\FOR{\texttt{id} $\gets$ 0 \textbf{to} \texttt{layer\_num}}
    \STATE \texttt{gain} $\gets$ \texttt{deviations[id]} - \\~~~~\texttt{deviations[id+1]}
    \STATE \texttt{layer\_gains.add(gain)}
\ENDFOR

\STATE \emph{\# (c) freeze layers and backpropagate}
\STATE \texttt{eof\_layer} $\gets$ \texttt{argmax(layer\_gains)}
\STATE \texttt{freeze\_layers(range(eof\_layer))}
\STATE \texttt{backpropagate(model, datum)}

\end{algorithmic}
\end{algorithm}


For a given dataset, the computation cost of backpropagation is decided by the depth of eof-layers, we can count the depths to know the cost-saving of layer-freezing. On the opposite, we can have a budget plan and force the depths of eof-layers to fulfill the budget. In this way, we control the cost-saving by planning the depth of eof-layers.

\subsection{Budget for Layer-Freezing}


To balance the effectiveness and cost of model fine-tuning, we incorporate a budget to determine the extent of layer-freezing based on specific requirements (see \cref{appendix:details}).
This budget represents the number of model layers to fine-tune for a given dataset. 
It controls the efficiency of LM finetuning, for example, we tend to give a low budget for LM finetuning if we want a high efficiency.

\paragraph{Budget Plan}
Similar to the common practice of finetuning half layers, we design budget plans to control the cost-benefit tradeoff.
For a given model of $m$ layers, we make the amount of data, that is assigned to finetuning layers between the \texttt{eof\_layer} to the last layer, following the relative proportion of the growth sequence:
(1) Following geometric growth, we take the growth ratio as $2$. Then, the amount of data assigned for finetuning follows the relative proportion of $1, 2, 4, ..., 2^{m-1}$;
(2) Following arithmetic growth, we make the initial term the common difference between terms. Then, the amount of data assigned for finetuning follows the relative proportion of $1, 2, 3, ..., m-1$.

\begin{figure}[!htb]
    \centering
    \includegraphics[width=1.0\linewidth]{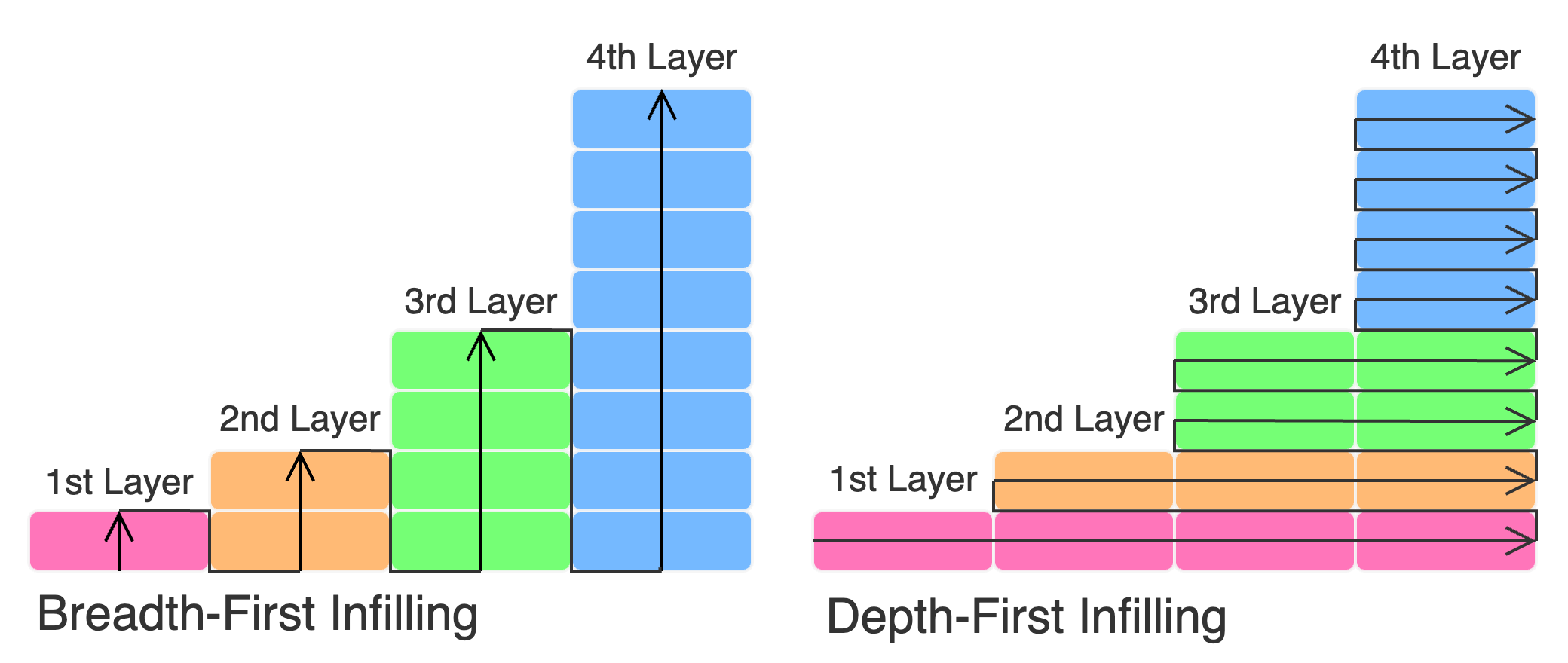}
    \caption{The order of budget infilling for a 4-layer model is: first red, then orange, then green, and finally blue shares.
 In breadth-first infilling, the color of shares is decided by layer. First let eof-layers be in first-layer until the layer is full (red); then let them be in second-layer until full (orange); then be in third-layer (green); and finally be in last-layer (blue).
 In depth-first infilling, the color of shares is decided by the position in layers. First let eof-layers be in first-share of all layers, from shallow to deep layers; then let them be in second-share of all layers; and then repeat the practice in the third share, forth share, until the budget of each layer is satisfied in the proper order (red, orange, green, and blue).
 }
    \label{fig:salf-budget}
\end{figure}

\paragraph{Budget Infilling}
For a given dataset, if the budget cannot be infilled completely with the data, the infilling order will affect the cost-benefit tradeoff.
We introduce two practices for budget infilling:
(1) Breadth-First (BF) fills eof-layers in the shallow layers, and then deeper layers;
(2) Depth-First (DF) fills eof-layers in all layers evenly, until layers are infilled successively from shallow to deep.
We illustrated with a model having four layers, following geometric growth, as shown in \cref{fig:salf-budget}.

\section{Experiments and Results}
\label{sec:experiments}




\subsection{Setup}

\paragraph{Datasets}
We use 5 established datasets, covering the common natural language tasks:
emotion recognition: CARER~\cite{Saravia2018CARERCA};
similarity detection: MRPC~\cite{Dolan2005AutomaticallyCA};
sentiment analysis: SST5~\cite{Socher2013RecursiveDM};
and general text classification: TREC~\cite{Voorhees2000BuildingAQ} and WebSS~\cite{Phan2008LearningTC}.
The statistics of datasets are shown in \cref{tab:corpus_stats}.

\begin{table}[!tb]
    \centering
    \resizebox{1.0\linewidth}{!}{%
        \sisetup{table-format=2.2}
\rowcolors{2}{}{gray!10}
\begin{tabular}{
    ll cccccc
}

\hiderowcolors
\toprule

& & \textbf{CARER} & \textbf{MRPC} & \textbf{SST5} & \textbf{TREC} & \textbf{WebSS} \\

\midrule
\multicolumn{2}{c}{Class Num.} & 6 & 2 & 5 & 6 & 8 \\

\midrule
\multirow{2}*{Data Num.} & Train & 16,000 & 4,076 & 8,544 & 5,452 & 10,060 \\
& Test & 2,000 & 1,725 & 2,210 & 500 & 2,280 \\

\midrule
\multicolumn{2}{c}{Avg. Prompt Length}
& 25.6 & 61.0 & 28.0 & 17.1 & 27.8 \\

\bottomrule

\end{tabular}

 }
    \caption{Stats of natural language datasets.}
    \label{tab:corpus_stats}
\end{table}

\paragraph{Models}
We use the recently released LLMs, including Qwen2 (0.5B-7B)~\cite{Yang2024Qwen2TR}, Gemma2 (2B-9B)~\cite{Riviere2024Gemma2I}, and the state-of-the-art Llama-3 (8.0B)~\footnote{https://github.com/meta-llama/llama3}. They are performant in massive comparisons with other competitors, and leading in the popularity statistics
(especially, most downloads per month) in the hugging-face website \footnote{https://huggingface.co/}.
The details are available in \cref{tab:model_stats}.

\begin{table}[!tb]
    \centering
    \resizebox{1.0\linewidth}{!}{%
        \sisetup{table-format=2.2}
\rowcolors{2}{}{gray!10}
\begin{tabular}{
    l ccccc c
}

\hiderowcolors
\toprule

& \multicolumn{3}{c}{\textbf{Qwen2}} & \multicolumn{2}{c}{\textbf{Gemma2}} & \multicolumn{1}{c}{\textbf{Llama3}} \\
\cmidrule(lr){2-4} \cmidrule(lr){5-6} \cmidrule(lr){7-7}
& {\textbf{0.5B}} & {\textbf{1.5B}} & {\textbf{7B}} & {\textbf{2B}} & {\textbf{9B}} & {\textbf{8B}} \\

\midrule

Model Size & 0.49B & 1.54B & 7.62B & 2.61B & 9.24B & 8.03B \\

\midrule

Head Num. & 14 & 12 & 28 & 8 & 16 & 32 \\

\midrule

Layer Num. & 24 & 28 & 28 & 26 & 42 & 32 \\

\midrule

Dimension & 896 & 1,536 & 3,584 & 2,304 & 3,584 & 4,096 \\

\midrule

Vocabulary & \multicolumn{2}{c}{151,936} & 152,064 & \multicolumn{2}{c}{256,000} & 128,256 \\

\bottomrule

\end{tabular}

 }
    \caption{Stats of Qwen2, Gemma2, and Llama3 models.}
    \label{tab:model_stats}
\end{table}

\paragraph{Baselines}
\textsc{LIFT} is the state-of-the-art in layer-wise LM finetuning on saving the computation cost. It takes a front-to-end selection policy to prioritize the layer to finetune~\cite{zhu2024lift}.
However, it only finetunes one layer each time, which may damage its performance.
We relax its restrictions for a stronger baseline by letting more layers be trainable while the computation cost is the same. We mark the vanilla one as \mytt{LIFT}{half}, and the enhanced one as \mytt{LIFT$\star$}{half}.
In addition, we also compare our approach \textsc{SALF} with two common finetuning practices with LoRA: full-layer finetuning and half-layer finetuning. The former is to finetune all model layers, while the latter is to finetune only the last half model layers.
We marked them as \mytt{LoRA}{full} and \mytt{LoRA}{half}.

\paragraph{Metrics}
For effectiveness, we use \emph{F1 score} to measure whether the predicted next-token is the ground truth because of the class imbalance in the datasets.
F1 score is the harmonic mean of precision and recall, and considers the effects of both false positives and false negatives.
For efficiency, we use the ``cost-saving'' ratio as a new metric, representing the saved computation cost in backpropagation. Large ratios mean better effects.

\paragraph{Pipeline}
We conduct LM finetuning experiments to compare our approach with other layer-freezing practices.
Since \textsc{LIFT} is designed to save around $50\%$ computation cost in backpropagation, we restrict our approach to the same computation cost for a fair comparison (on effectiveness).
Our implementation is as we described, and does not use other techniques, such as instruction tuning.
The details on the implementation are in \cref{appendix:details}.

\subsection{Performance Evaluation}

We evaluate the performance of our approach and the baselines in LM finetuning: finetune the LMs on the training set, and do inference on the test set.

\begin{table}[!tb]
    \centering
    \resizebox{1.0\linewidth}{!}{%
        \sisetup{table-format=2.2}
\begin{tabular}{
    cl ccccc c
}

\hiderowcolors
\toprule

\multirow{2}{*}{\textbf{LLM}} & \multirow{2}{*}{\textbf{Method}}
& \multicolumn{5}{c}{\textbf{Dataset}} & \multirow{2}{*}{\textbf{Avg.}} \\
\cmidrule(lr){3-7}
& & \textbf{CARER} & \textbf{MRPC} & \textbf{SST5} & \textbf{TREC} & \textbf{WebSS} & \\

\midrule
\multirow{5}{*}{\rotatebox[origin=c]{90}{Qwen2-0.5B}}
& \mytt{LoRA}{full} & 0.765 & 0.454 & 0.302 & 0.779 & 0.837 & 0.627 \\
& \mytt{LoRA}{half} & 0.746 & 0.663 & 0.335 & 0.795 & 0.847 & 0.677 \\
& \mytt{LIFT}{half} & 0.241 & 0.399 & 0.239 & 0.701 & 0.583 & 0.433 \\
& \mytt{LIFT$\star$}{half} & 0.806 & 0.755 & \tabh 0.456 & 0.805 & \tabh 0.891 & 0.743 \\
& \mytt{\textbf{SALF}}{half} & \tabh 0.807 & \tabh 0.785 & 0.444 & \tabh 0.941 & 0.847 & \tabh 0.765 \\

\midrule
\multirow{5}{*}{\rotatebox[origin=c]{90}{Qwen2-1.5B}}
& \mytt{LoRA}{full} & \tabh 0.835 & 0.750 & 0.293 & 0.787 & 0.749 & 0.683 \\
& \mytt{LoRA}{half} & 0.687 & 0.769 & 0.373 & 0.793 & 0.856 & 0.696 \\
& \mytt{LIFT}{half} & 0.469 & 0.566 & 0.318 & 0.727 & 0.705 & 0.557 \\
& \mytt{LIFT$\star$}{half} & 0.823 & \tabh 0.779 & 0.503 & 0.808 & \tabh 0.917 & 0.766 \\
& \mytt{\textbf{SALF}}{half} & 0.815 & 0.735 & \tabh 0.520 & \tabh 0.940 & 0.876 & \tabh 0.777 \\

\midrule
\multirow{5}{*}{\rotatebox[origin=c]{90}{Qwen2-7B}}
& \mytt{LoRA}{full} & 0.820 & 0.399 & 0.075 & 0.252 & 0.029 & 0.315 \\
& \mytt{LoRA}{half} & 0.794 & 0.690 & 0.388 & 0.796 & 0.866 & 0.707 \\
& \mytt{LIFT}{half} & 0.532 & 0.739 & 0.320 & 0.747 & 0.699 & 0.607 \\
& \mytt{LIFT$\star$}{half} & 0.797 & 0.781 & 0.534 & 0.802 & \tabh 0.915 & 0.766 \\
& \mytt{\textbf{SALF}}{half} & \tabh 0.823 & \tabh 0.831 & \tabh 0.542 & \tabh 0.951 & 0.852 & \tabh 0.800 \\

\midrule
\multirow{5}{*}{\rotatebox[origin=c]{90}{Gemma2-2B}}
& \mytt{LoRA}{full} & 0.867 & 0.494 & \tabh 0.281 & 0.711 & 0.771 & 0.625 \\
& \mytt{LoRA}{half} & 0.865 & 0.498 & 0.243 & 0.711 & 0.694 & 0.602 \\
& \mytt{LIFT}{half} & 0.328 & 0.399 & 0.082 & 0.505 & 0.453 & 0.353 \\
& \mytt{LIFT$\star$}{half} & 0.872 & \tabh 0.518 & 0.280 & \tabh 0.737 & \tabh 0.797 & \tabh 0.641 \\
& \mytt{\textbf{SALF}}{half} & \tabh 0.877 & 0.399 & 0.199 & 0.734 & 0.778 & 0.597 \\

\midrule
\multirow{5}{*}{\rotatebox[origin=c]{90}{Gemma2-9B}}
& \mytt{LoRA}{full} & 0.801 & \tabh 0.399 & 0.193 & 0.706 & 0.765 & 0.573 \\
& \mytt{LoRA}{half} & \tabh 0.865 & \tabh 0.399 & 0.201 & 0.725 & 0.741 & 0.586 \\
& \mytt{LIFT}{half} & 0.382 & \tabh 0.399 & 0.187 & 0.577 & 0.484 & 0.406 \\
& \mytt{LIFT$\star$}{half} & 0.862 & \tabh 0.399 & \tabh 0.281 & \tabh 0.729 & 0.791 & \tabh 0.612 \\
& \mytt{\textbf{SALF}}{half} & 0.860 & \tabh 0.399 & 0.190 & 0.658 & \tabh 0.797 & 0.581 \\

\midrule
\multirow{5}{*}{\rotatebox[origin=c]{90}{Llama3-8B}}
& \mytt{LoRA}{full} & 0.394 & 0.399 & 0.402 & 0.256 & 0.029 & 0.296 \\
& \mytt{LoRA}{half} & 0.818 & \tabh 0.664 & 0.338 & 0.784 & 0.861 & 0.693 \\
& \mytt{LIFT}{half} & 0.590 & 0.466 & 0.476 & 0.779 & 0.837 & 0.630 \\
& \mytt{LIFT$\star$}{half} & 0.843 & 0.468 & 0.552 & 0.799 & \tabh 0.900 & 0.712 \\
& \mytt{\textbf{SALF}}{half} & \tabh 0.872 & 0.399 & \tabh 0.571 & \tabh 0.945 & 0.885 & \tabh 0.734 \\

\bottomrule

\end{tabular}

 }
    \caption{F1 Scores of layer-freezing methods (on the diverse datasets and models).}
    \label{tab:results_q1_measure}
\end{table}

As shown in \cref{tab:results_q1_measure}, based on the average F1 score, on 4 out of 6 models, \textsc{SALF} performs better than others, while on the other model, its performance is very close to the best.
Compared with the common practices \mytt{LoRA}{full} and \mytt{LoRA}{half}, \textsc{LIFT} shows superiority in the performance while our approach \textsc{SALF} shows stable and obvious improvements.
Besides, the advantages of \textsc{SALF} vary on the datasets. On WebSS, \textsc{SALF} performs the best only in the case where the model is Llama3, but the performance gap to the best is not obvious. However, \textsc{SALF} cannot show stable improvements on MRPC, especially when with Gemma2 and Llama3.
The reason is that, the class number of the MRPC dataset is only $2$, meaning the semantic transition is very simple, thereby the deviations in the process may not be very helpful.
All methods cannot perform well in SST5 with Gemma2, which may caused by the bad semantic property of Gemma2 models due to its risk of semantic degradation in Gemma2 design.
Gemma2 models use a non-standard FFN, which has an extra gate on top of GEGLU. In contrast, Qwen2 and Llama3 models use SwiGLU, which is a standard FFN design. Another possible reason is that, Gemma2 uses SentencePiece as the tokenizer, which is not as good as the BPE tokenizer used in Qwen2 and Llama3, especially in token granularity and consistency.

It is noteworthy that \mytt{LoRA}{full} performs worse than others, and even worse than \mytt{LoRA}{half}.
It is counter-intuitive since full-layer finetuning is updating all layers and requires a larger computation cost than \mytt{LoRA}{half}.
However, in our understanding, it may caused by the difference in the effects of shallow and deep model layers. Usually, shallow layers learn the macro features while deep layers learn the micro features~\cite{So2019TheET,Brown2020LanguageMA}.
It means, \emph{when the learning rate is fixed in LM finetuning, the update in shallow layers shall be less frequent than that in deep layers}.
It also explains the reason why both \mytt{LoRA}{full} and \mytt{LoRA}{half} perform not as well as \textsc{LIFT$\star$} or \textsc{SALF}: \mytt{LoRA}{full} updates shallow layers too often while \mytt{LoRA}{half} updates them too seldom.

Meanwhile, the results of the baseline \textsc{LIFT} is not as good as the enhanced implementation \textsc{LIFT$\star$}. Their difference is that, the former only makes the eof-layer trainable, while the latter finetunes all layers between eof-layer to last-layer.
It indicates that, merely finetuning shallow layers cannot guarantee smaller deviations in the deep layers, or the deviations require further processing.

In \cref{appendix:analysis}, we compared the effects of taking different metrics of deviations.
Further, we analyzed the advantages of \textsc{SALF} in LM finetuning with the illustrations on semantic deviations.


\section{Analysis on Cost-Benefit Tradeoff}
\label{sec:analysis}


We study the performance and cost-benefit tradeoff of budget plans and infilling practices to layer-freezing.
For example, a geometric-growth budget with breadth-first infilling is denoted as \mytt{geom}{bf}.

As shown in \cref{tab:results_q3_budget}, the budget for cost-benefit tradeoff is useful to both \textsc{LIFT$\star$} and our approach \textsc{SALF}, while our approach still shows better performance.
In comparison, the arithmetic-growth budget shows similar performance to the geometric-growth budget.
Meanwhile, the practice of depth-first infilling tends to perform better and more stably than breadth-first infilling.


\begin{table}[!htb]
    \centering
    \resizebox{1.0\linewidth}{!}{%
        \sisetup{table-format=2.2}
\begin{tabular}{
    l ccccc c
}

\hiderowcolors
\toprule

\multirow{2}{*}{\textbf{Budget}}
& \multicolumn{5}{c}{\textbf{Dataset}} & \multirow{2}{*}{\textbf{Avg.}} \\
\cmidrule(lr){2-6}
& \textbf{CARER} & \textbf{MRPC} & \textbf{SST5} & \textbf{TREC} & \textbf{WebSS} & \\

\midrule
\mytt{LIFT$\star$}{half} & 0.843 & 0.468 & 0.552 & \tabh 0.799 & 0.900 & 0.712 \\
\myttt{LIFT$\star$}{arith}{bf} & 0.817 & 0.516 & 0.548 & 0.798 & 0.893 & 0.714 \\
\myttt{LIFT$\star$}{arith}{df} & 0.835 & 0.581 & 0.543 & 0.789 & \tabh 0.906 & 0.731 \\
\myttt{LIFT$\star$}{geom}{bf} & 0.763 & \tabh 0.729 & 0.404 & 0.791 & 0.876 & 0.713 \\
\myttt{LIFT$\star$}{geom}{df} & \tabh 0.845 & 0.625 & \tabh 0.559 & 0.795 & 0.899 & \tabh 0.745 \\

\midrule
\mytt{\textbf{SALF}}{half} & 0.872 & 0.399 & 0.571 & 0.945 & 0.885 & 0.734 \\
\myttt{\textbf{SALF}}{geom}{bf} & 0.906 & 0.665 & 0.586 & 0.962 & 0.855 & 0.795 \\
\myttt{\textbf{SALF}}{geom}{df} & 0.920 & 0.711 & \tabh 0.607 & \tabh 0.970 & \tabh 0.921 & \tabh 0.826 \\
\myttt{\textbf{SALF}}{arith}{bf} & \tabh 0.921 & \tabh 0.752 & 0.391 & 0.964 & 0.911 & 0.788 \\
\myttt{\textbf{SALF}}{arith}{df} & 0.914 & 0.751 & 0.588 & 0.964 & 0.914 & \tabh 0.826 \\

\bottomrule

\end{tabular}

 }
    \caption{Accuracy of layer-freezing methods with different budget plans and infilling practices (on the diverse datasets, using Llama3-8B).}
    \label{tab:results_q3_budget}
\end{table}


As shown in \cref{tab:results_q3_cost}, compared with geometric-growth, the budget of arithmetic-growth saves more computation costs.
The reason is that, for a model of the same number of layers, the arithmetic-growth increases slower than the geometric-growth, so the budget of the latter is not likely to be fulfilled. For geometric-growth, eof-layers can fill in shallow layers but cannot fill in deep layers.
Also, depth-first infilling saves more than the breadth-first infilling. The reason is similar, more eof-layers tend to be in deep layers than in shallow layers.

\begin{table}[!htb]
    \centering
    \resizebox{1.0\linewidth}{!}{%
        \sisetup{table-format=2.2}
\begin{tabular}{
    l ccccc c
}

\hiderowcolors
\toprule

\multirow{2}{*}{\textbf{Budget}}
& \multicolumn{5}{c}{\textbf{Dataset}} & \multirow{2}{*}{\textbf{Avg.}} \\
\cmidrule(lr){2-6}
& \textbf{CARER} & \textbf{MRPC} & \textbf{SST5} & \textbf{TREC} & \textbf{WebSS} & \\

\midrule
\mytt{LoRA}{full} & 0.000 & 0.000 & 0.000 & 0.000 & 0.000 & 0.000 \\
\mytt{LoRA}{half} & \tabh 0.500 & \tabh 0.500 & \tabh 0.500 & \tabh 0.500 & \tabh 0.500 & \tabh 0.500 \\
\mytt{LIFT}{half} & 0.484 & 0.483 & 0.484 & 0.484 & 0.484 & 0.484 \\
\mytt{LIFT$\star$}{half} & 0.484 & 0.483 & 0.484 & 0.484 & 0.484 & 0.484 \\
\mytt{\textbf{SALF}}{half} & 0.484 & 0.483 & 0.484 & 0.484 & 0.484 & 0.484 \\

\midrule
\myttt{}{geom}{bf} & 0.374 & 0.312 & 0.346 & 0.328 & 0.355 & 0.343 \\
\myttt{}{geom}{df} & 0.616 & 0.583 & 0.601 & 0.589 & 0.604 & 0.598 \\
\myttt{}{arith}{bf} & 0.614 & 0.613 & 0.613 & 0.609 & 0.615 & 0.613 \\
\myttt{}{arith}{df} & \tabh 0.644 & \tabh 0.640 & \tabh 0.644 & \tabh 0.642 & \tabh 0.645 & \tabh 0.643 \\

\bottomrule

\end{tabular}

 }
    \caption{Backpropagation cost-saving of layer-freezing methods with different budget plans (on the diverse datasets, using Llama3-8B).}
    \label{tab:results_q3_cost}
\end{table}

Considering the efficiency and cost-benefit tradeoff, the budget of arithmetic-growth shows equivalent performance but saves more computation costs. Also, the practice of depth-first infilling is better than breadth-first infilling.
Based on the results, an arithmetic-growth with depth-first infilling saves around $1/3$ more computation cost and has a slightly better performance.
The reason explaining why the combination is performant is the same as discussed, \emph{when the learning rate is fixed in LM finetuning, the update in shallow layers shall be less frequent than that in deep layers}.

\section{Related Work}
\label{sec:related_work}


Leveraging the layered structure of neural models, the concept of layer-freezing was proposed decades ago, but mainly for deep belief networks (DBN)~\cite{Hinton2009DeepBN}.
DBN is a stack of directed sigmoid belief network (SBN)~\cite{Neal1992ConnectionistLO} and an undirected restricted boltzmann machine~\cite{Hinton2017BoltzmannM}. The backpropagation is only applied to finetune the restricted boltzmann machine, while the dependencies between other layers are not bidirectional.
Therefore, progressively training each layer is proposed as a greedy strategy for training DBN~\cite{Hinton2006AFL,Bengio2006GreedyLT}.

In the era of language models, there has been little significant work studying layer-freezing for efficient finetuning, while the focus often lies on parameter-efficient, namely reducing the amount of trainable parameters, instead of computation-efficicent~\cite{Pan2024LISALI,zhu2024lift}.
One reason is the complexity and interpretability of language models. Besides, the correlation between model layers is not intuitive, and the effects of bidirectional dependencies on layer-wise finetuning have not been studied.
Another reason is that, the prior work on PEFT shows similar effects on reducing the number of trainable parameters, or even making the trainable parameters detachable.

\section{Conclusion}
\label{sec:conclusion}

In this paper, we have proposed the novel concept of semantic transition. By defining transition trace to describe the change of semantic meaning of the next token, we explain LM finetuning as the process of letting the representation gradually steer to the corresponding ground truth in latent space.
Meanwhile, based on a derived law of scaling law,
we can reasonably estimate and compare the capability of model layers, so to better allocate the computation resources in LM finetuning.
Further, we propose layer-freezing to accelerate LM finetuning, by finding the layer with the maximum gains of reducing deviation and finetuning deeper layers.

Based on our results on diverse datasets and multiple models, semantic-aware layer-freezing provides better performance than the state-of-the-art as well as common practices.
Moreover, our work explores the effects of budget plans on the cost-benefit tradeoff for layer-freezing.
%
In return, the effectiveness of our lay-finetuning approach validates the usefulness of semantic transition.


\section*{Limitations}
In this paper, we proposed semantic transition as a new perspective on the LMs' functionality. Besides, estimated and compared the capability of model layers. We suggest using the gains of reducing deviations in semantic transition to reduce the computation cost of LM finetuning. Our approach maintains and even improves the performance of LM finetuning.

In our understanding, our approach is leveraging the derived formula of scaling law to estimate and compare the capability of model layers.
However, the capability cannot be strictly seen as the convergence degree, namely the expected benefits of finetuning a certain model layer.
Besides, freezing the layer with the maximum gains of reducing deviation and finetuning shallower layers is an empirical wise practice, but there is no proof saying this is optimal.
Meanwhile, in a high-dimensional latent space, the representations tend to be orthogonal to each other~\cite{Vershynin_2018}. Therefore, using the cosine distance between latent representation and the semantic basis as the deviation may not be optimal practice. There possibly exists potential evidence to support other better choices.

The semantic transition is based on the similarity measurement between latent representations and semantic bases. The theoretical support is the local isotropy of LM latent space~\cite{Cai2021IsotropyIT}, therefore for the language models whose latent space cannot fulfill local isotropy in terms of semantics (even though they seem not to exist, to the best of our knowledge), our approach may not stand.

\bibliography{custom}

\appendix

\section{Implementation Details}
\label{appendix:details}

\subsection{Environments}

Our implementation uses deep learning framework \textsc{PyTorch}~\cite{Paszke2019PyTorchAI}, \textsc{Transformers}~\cite{Wolf2019TransformersSN}, and use \textsc{PEFT}~\footnote{https://github.com/huggingface/peft} to conduct the LoRA experiments.
The LM finetuning experiments are based on existing PEFT methods, specifically \textsc{LoRA}~\cite{Hu2021LoRALA}.
We use quantization techniques (INT4) to load Qwen2-7B, Gemma2-9B, and Llama3-8B, with the default settings~\cite{Dettmers2023QLoRAEF,Wu2023UnderstandingIQ}, which reduces the memory requirements in LM finetuning with slight performance loss.

The experiments are conducted via a single run, with the global random-seed $42$.
The computation is based on a single Nvidia V100 (\num{32} \text{GB}), and the computation budget is around 2200 GPU hours.

\subsection{License and Terms}

We understand and respect the licenses used in our experiments, including the Apache-2.0 license for Qwen2 models and Gemma2 models, as well as the Llama3 community license for Llama3 models \footnote{https://llama.meta.com/llama3/license/}.
We confirm that our use of existing artifacts was consistent with their intended use.




\subsection{SALF Algorithm with Budget}

By introducing the budget for LM finetuning, our semantic-based layer-freezing approach can fulfill the intended computation cost.
Then, to guarantee improved performance, we propose the \textsc{SALF} algorithm with the budget consideration, as shown in \cref{algo:salf-budget}.

\begin{algorithm}
\caption{SALF w/ Budgets}
\label{algo:salf-budget}
\begin{algorithmic}[1]

\REQUIRE \texttt{model}, \texttt{data}, \texttt{budgets}

\STATE \texttt{tabu\_data} $\gets$ \texttt{empty list}

\FOR{\texttt{layer} $\gets$ 0 \textbf{to} \texttt{layer\_num}}
    \STATE \emph{\# (a) freeze layers from deep to shallow}
    \STATE \texttt{freeze\_layers(range(layer))}

    \STATE \emph{\# Backpropagation of Matching Data}
    \FOR{\texttt{datum} \textbf{in} \texttt{data}}
        \STATE \emph{\# (b) check whether to jump the loop}
        \IF{\texttt{budgets[layer]} == 0}
            \STATE \textbf{break}
        \ENDIF
        \IF{\texttt{datum} \textbf{in} \texttt{tabu\_data}}
            \STATE \textbf{continue}
        \ENDIF

        \STATE \emph{\# (c) execute line 1-16 in \cref{algo:salf}}
        \STATE \texttt{eof\_layer} $\gets$ \texttt{SALF(model, datum)}
        \IF{\texttt{eof\_layer} > \texttt{layer}}
            \STATE \textbf{continue}
        \ENDIF

        \STATE \emph{\# (d) backpropagate}
        \STATE \texttt{backpropagate(model, datum)}
        \STATE \texttt{budgets[layer] -= 1}
        \STATE \texttt{tabu\_data.append(datum)}
    \ENDFOR

    \STATE \emph{\# Backpropagation of Remaining Data}
    \STATE \texttt{sampled\_data} $\gets$ \texttt{random\_sample(\\~~data, filter=tabu\_data, \\~~amount=budgets[layer])}
    \STATE \texttt{finetune(model, sampled\_data)}
    \STATE \texttt{budgets[layer]} $\gets$ \texttt{0}
    \STATE \texttt{tabu\_data.extend(sampled\_data)}
\ENDFOR

\end{algorithmic}
\end{algorithm}

The intent of the code is intuitive: first, compute the deviations to find the eof-layer for each data; then, arrange the data with the similar eof-layers into the budget; last, gradually narrow down the scope of finetuning (freezing more model layers), and use the arranged data to backpropagate the loss.
%
For the sake of the sequential access restriction of \texttt{data-loader}, the algorithm is described with the for-loops and the repeated iterations.
In the implementation, we can use random access and caching techniques to remove the for-loops and reduce the number of iterations.


\section{More Analysis}
\label{appendix:analysis}

\subsection{Differences between Parameter-Efficiency and Computation-Efficiency}

Different from PEFT methods proposed for better parameter-efficiency, our approach \textsc{SALF} (as well as the baseline \textsc{LIFT}) is a layer-freezing method proposed for better computation-efficiency.
The focus of parameter-efficiency is reducing the memory cost of finetuning, while in contrast, the focus of computation-efficiency is reducing the computation cost of backpropagation.
For newly-emerging topics, including knowledge editing~\cite{Yao2023EditingLL}, representation engineering~\cite{Zou2023RepresentationEA}, and language model repair~\cite{Gu2023NeuronPS}, computation-efficiency is critical in realizing the flexibility and adaptability.

Compared with full-parameter finetuning, PEFT methods cannot guarantee computation-efficiency.
The computation cost of finetuning covers the cost of forward-inference and back-propagation. The forward-inference cost cannot be reduced, so any method for better computation-efficiency must deal with the back-propagation cost. Then, based on the \emph{chain rule} of calculus to compute gradients, which is the mathematical foundation of back-propagation, if the gradients of the $k$-th layer are needed, the gradient computation of any deeper layers (whose layer index is larger than $k$) cannot be skipped. Therefore, PEFT methods like \textsc{LoRA} are not computation-efficient since they cannot reduce the cost of back-propagation.
For the same reason, layer-freezing is intuitive and reliable in guaranteeing the computation-efficiency.

\subsection{Metrics for Computing Deviations}

\textsc{SALF} represents a common practice to detect how the model capability improves across different layers. That is, probing the hidden states in middle layers, and using them for logits computation as an estimation for LM interpretability.

When computing the deviations in LM inference, there are alternatives to the used semantic cosine-distance loss.
We check the case where letting the cross-entropy loss be the deviation measurement, denoted as \myttt{SALF}{half}{ce}.
Since cross-entropy loss is do computation with all ground truths, not merely with the corresponding one, as did by cosine-distance loss, the former one involves more constraints than the latter one. It indicates that \myttt{SALF}{half}{ce} will be slower in convergence, and further explains why this variant cannot perform as well as \mytt{SALF}{half} when training for the same epoch. Based on our analysis, they tend to have similar performance when doing model finetuning for an unlimited number of epochs until convergence.
In our understanding, \emph{a less constrained loss function indicates a more straightforward convergence process, and therefore tends to perform better in LM finetuning}.
Since the cross-entropy loss is commonly used in logits computation, the advantages of \textsc{SALF} indicate that, cosine-distance loss is a notable alternative for its better efficiency.

\begin{table}[!tb]
    \centering
    \resizebox{1.0\linewidth}{!}{%
        \sisetup{table-format=2.2}
\begin{tabular}{
    l ccccc c
}

\hiderowcolors
\toprule

\multirow{2}{*}{\textbf{Variant}}
& \multicolumn{5}{c}{\textbf{Dataset}} & \multirow{2}{*}{\textbf{Avg.}} \\
\cmidrule(lr){2-6}
& \textbf{CARER} & \textbf{MRPC} & \textbf{SST5} & \textbf{TREC} & \textbf{WebSS} & \\


\midrule
\mytt{\textbf{SALF}}{half} & \tabh 0.872 & 0.399 & \tabh 0.571 & 0.945 & \tabh 0.885 & \tabh 0.734 \\
\myttt{\textbf{SALF}}{half}{ce} & 0.213 & 0.595 & 0.419 & \tabh 0.952 & 0.835 & 0.603 \\
\myttt{\textbf{SALF}}{half}{rank} & 0.086 & \tabh 0.753 & 0.455 & 0.946 & 0.876 & 0.623 \\

\bottomrule

\end{tabular}

 }
    \caption{F1 scores of layer-freezing variants (on the diverse datasets, using Llama3-8B).}
    \label{tab:results_q2_ablation}
\end{table}


Meanwhile, we experimented with a variant using the customized metric: \myttt{SALF}{half}{rank} measures the ranking of the ground truth in the output probabilities. Theoretically, in LM finetuning, the ranking of the ground truth shall keep increasing until becoming the first.
As shown in \cref{tab:results_q2_ablation}, it fails to realize the equivalent performances to \textsc{SALF}. Based on our analaysis, it is caused by the small output space and the large model size. For example, Llama3-8B has 32 model layers while the class number of datasets is smaller than 10, so the deviations tend to be very small, and so do the gains in reducing the deviations. The variant \myttt{SALF}{half}{rank} cannot be numerically sensitive, since its deviations tend to remain unchanged in neighboring layers and the gains cannot express useful information.
In contrast, the cosine-distance loss is numerically sensitive, and focuses on cosine similarity with the corresponding ground truth.

\subsection{Heuristics in Layer Freezing}

Based on the analytical results on semantic transition in LMs, namely deviations and gains, there are heuristic practices to decide which layers to freeze.

In our approach, we find the model layer where the gain is the largest, and freeze it and other deeper layers. Concerning the effects, after multiple turns of layer-freezing, deeper layers are more likely to become the largest-gain layer, especially the deepest layer.
Considering the functionality of each LM layer contributes to the convergence monotonically~\cite{Belrose2023ElicitingLP}, our approach tends to force each layer contribute to the converge. Correspondingly, the gain of a deep layer will be larger than that of a shallow layer.

We conducted a contrastive study on other similar decisions, such as finding the least-deviation layer, and then freeze it and other deeper layers.
We observed that the located layer is usually the first model layer, because the first model layer usually improves the largest model capability, which means, the deviation gain is the least. Therefore, if the strategy of layer-freezing is to find the layer of the least gain, the effect is like freezing no layers.

As for the heuristic options which decide layer-freezing directly using deviations, we find their benefits to LM finetuning are not stable.
It indicates that, the comparisons in adjacent layers provide more useful information than directly using the deviations. Based on our understanding, layer-by-layer comparison serves the function of regularizing deviations. The necessity of regularization is because the layer-wise deviations correspond to different latent spaces, so the deviations cannot be compared across model layers.

\subsection{Semantic Effects of LM Finetuning}

\begin{figure*}[!htb]
    \centering
    \includegraphics[width=1.0\linewidth]{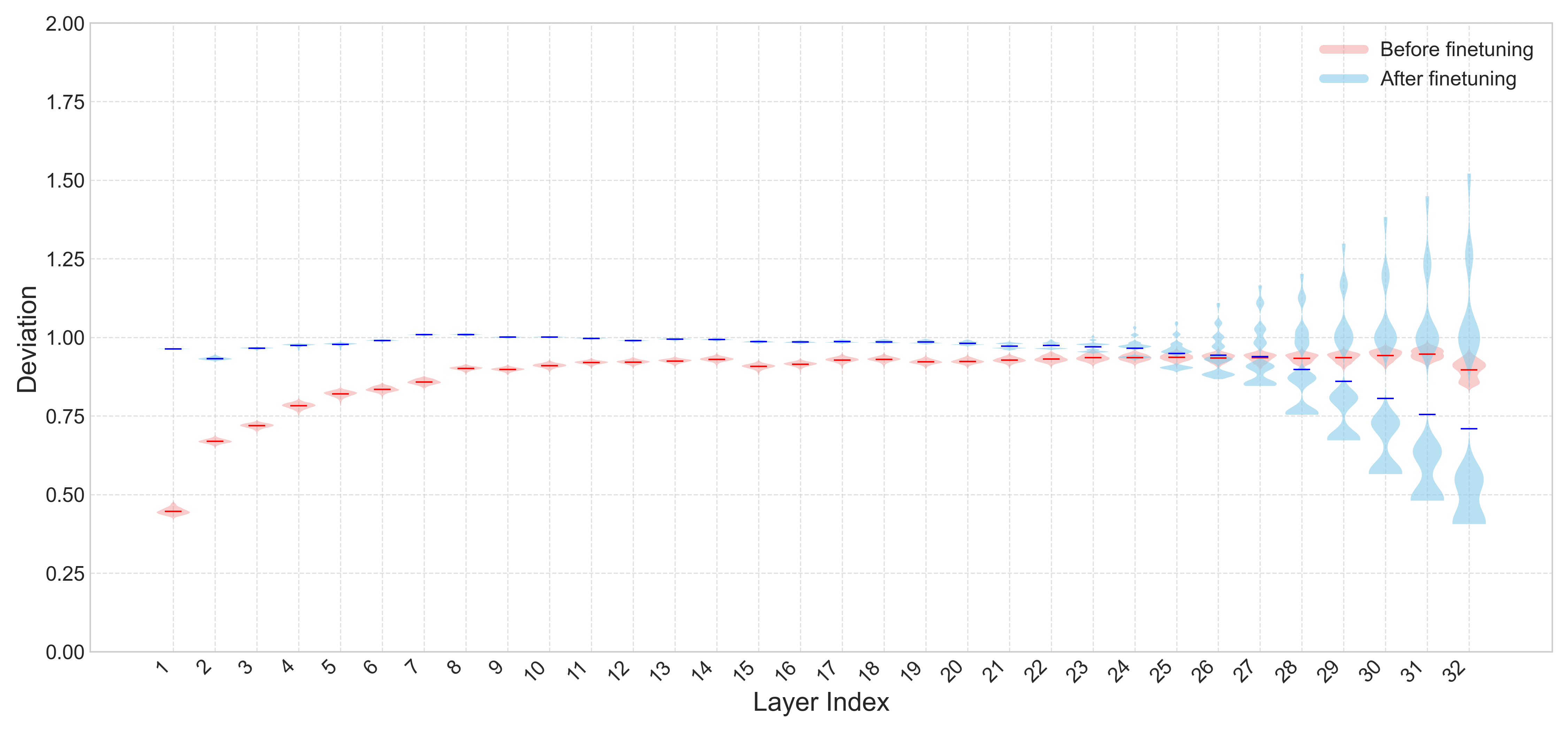}
    \caption{
 Violin plot of the deviations of each layer in LM finetuning, where the crossbars represent the mean of deviations, when making all model layers trainable (on the CARER dataset, using Llama3-8B).
 The phenomena include:
 (1) The red crossbars usually lie at lower positions than the blue crossbars (in the first 27 layers). It means, the deviation changes by LM finetuning are negative in most layers.
 (2) The blue shapes are flattened in the last few layers (from the $25$-th layer to the last-layer) but some areas in the shapes lie at higher positions. It means, the distribution of the deviations in the last layers is forming multiple peaks, no longer centered in only one peak, and lots of data show higher deviations;
 (3) The differences between red and blue are large and show a reversal (first red is better, then blue is better) in the first and last few layers. It means, the deviation changes by LM finetuning are significant, which are worse in the shallower layers but better in the deeper layers.
 }
    \label{fig:loss-none}
\end{figure*}

\begin{figure*}[!htb]
    \centering
    \includegraphics[width=1.0\linewidth]{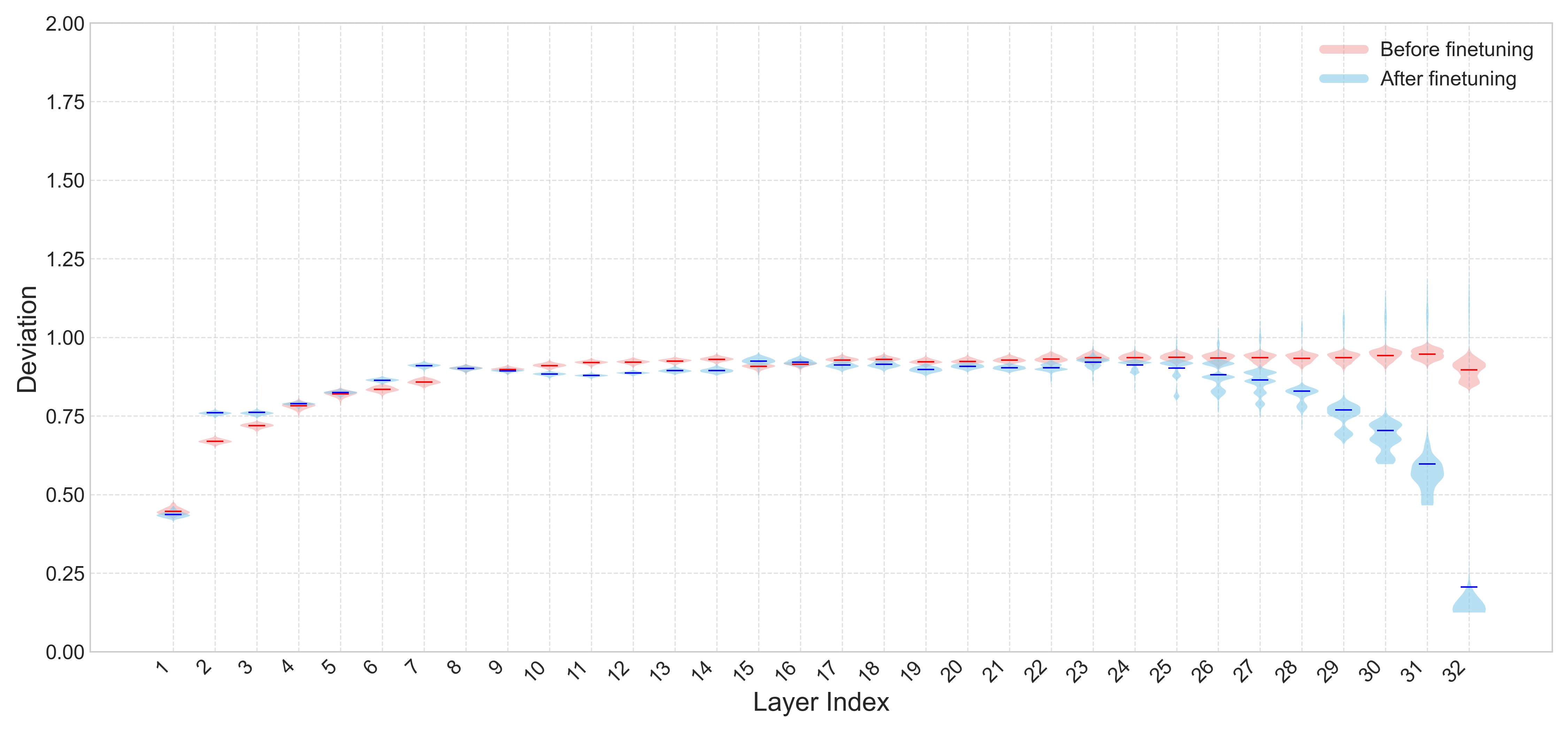}
    \caption{
 Violin plot of the deviations of each layer in LM finetuning, where the crossbars represent the mean of deviations, when taking semantic-based layer-freezing (on the CARER dataset, using Llama3-8B).
 The phenomena include:
 (1) The red crossbars usually lie at the same positions as the blue crossbars (in the first 27 layers). It means, the deviation changes by LM finetuning are very small in most layers.
 (2) The blue shapes are flattened in the last few layers (from the $25$-th layer to the last-layer) but almost all areas in the shapes lie at lower positions. It means, the distribution of the deviations in the last layers is forming multiple peaks, no longer centered in only one peak, and almost all data show lower deviations;
 (3) The differences between red and blue are only getting large (blue is better) in the last few layers. It means, the deviation changes by LM finetuning are positive and highly targeted, which are mainly in the deeper layers.
 }
    \label{fig:loss-laft}
\end{figure*}

To study the effects of our SALF approach on LM finetuning, we illustrate the deviation changes in LM finetuning of two settings:
one is making all layers trainable, corresponding to \mytt{LoRA}{full}, as shown in \cref{fig:loss-none}; while the other one is taking our approach for layer-freezing, corresponding to \mytt{SALF}{half}, as shown in \cref{fig:loss-laft}.

In the illustrations, the deviations are in the range of $[0, 2]$, since it is derived from the cosine similarity.
Besides, in a high-dimensional latent space, the representations tend to be orthogonal to others (including the semantic bases)~\cite{Vershynin_2018}, so when the deviations are smaller than $1$, it means the corresponding data representations are steering towards the ground truth, then the corresponding LM predictions may be correct. Otherwise, if the deviations are larger than $1$, then the corresponding LM predictions are not likely to be correct.

By comparing the illustrated two situations of the blue shapes, we conclude the advantages of our semantic-based layer-freezing approach to LM finetuning as:
our approach avoids the side effects of LM finetuning to shallow layers, and tends to make the semantic deviations in deep layers small.
Taking the illustrated situation of the red shapes as a reference, we believe that the first advantage (on the side effects to the shallow layers) may be the cause of the second advantage (on the small deviations in deep layers). It explains why our approach lead to small deviations in deep layers, and also, it also emphasizes the importance of reducing the deviations in shallow layers.
Further, the causation explains how to achieve better performance while reducing the computation cost in LM finetuning.

In addition, based on the illustrations, we see the accumulated effects of our approach in reducing the deviations in the last few model layers, where the blue shapes gradually move to lower positions, which indicates lower deviations of the data and a higher likelihood of correct LM predictions.

\end{document}